%% file: main.tex
\documentclass[10pt,twocolumn,letterpaper]{article}

\usepackage{cvpr}              

\input{preamble}

\definecolor{cvprblue}{rgb}{0.21,0.49,0.74}
\usepackage[pagebackref,breaklinks,colorlinks,citecolor=cvprblue]{hyperref}
\usepackage{adjustbox}
\usepackage{stfloats}
\usepackage{subcaption}
\usepackage{multirow}
\usepackage{makecell}
\usepackage{float}

\usepackage{stfloats}
\usepackage{amsmath,amsthm,amssymb,amsfonts}
\usepackage{mathtools}

\usepackage{MnSymbol}%
\usepackage{wasysym}%
\usepackage{multirow} 
\usepackage{xspace}

\theoremstyle{plain}

\theoremstyle{definition}

\theoremstyle{remark}

\usepackage{adjustbox}
\usepackage{graphicx}
\usepackage{dsfont}
\usepackage{tabulary}
\usepackage{tabularx}
\usepackage{makecell}
\usepackage{wrapfig}
\usepackage{fancyvrb}
\usepackage{MnSymbol}%
\usepackage{wasysym}%
\usepackage{multirow} 
\usepackage{xspace}

\usepackage{dsfont}
\newcommand{\EE}[0]{\mathbb{E}}

\newcommand{\set}[1]{\ensuremath{{#1}}}

\def\arrvline{\hfil\kern\arraycolsep\vline\kern-\arraycolsep\hfilneg}

\newcommand{\g}[1]{\mathcal{{#1}}}

\newcommand{\s}[1]{\mathcal{{#1}}}
 
\newcommand{\pr}[1]{\left(#1 \right)} 
\newcommand{\br}[1]{\left[#1 \right]} 

\renewcommand{\v}[1]{\ensuremath{\mathbf{#1}}} 

\newcommand{\argmin}{\operatornamewithlimits{arg\,min}}

\definecolor{subduedgreen}{RGB}{34, 139, 34} 

\def\paperID{*****} 
\def\confName{CVPR}
\def\confYear{2024}

\title{Improved Canonicalization for Model Agnostic Equivariance}

\author{Siba Smarak Panigrahi, Arnab Kumar Mondal\\
McGill University \& Mila\\
{\tt\small \{siba-smarak.panigrahi,arnab.mondal\}@mila.quebec}
}

\DeclareUnicodeCharacter{0301}{\'{e}}

\begin{document}
\maketitle
\input{sec/0_abstract}    
\input{sec/1_intro}

\input{sec/2_background}
\input{sec/3_method}
\input{sec/4_results}
\input{sec/5_discussion}
\input{sec/6_limitations_future_work}
{
    \small
    \bibliographystyle{ieeenat_fullname}
    \bibliography{main}
}


\end{document}

%% file: preamble.tex
\usepackage[dvipsnames]{xcolor}


%% file: sec/0_abstract.tex
\begin{abstract}

This work introduces a novel approach to achieving architecture-agnostic equivariance in deep learning, particularly addressing the limitations of traditional layerwise equivariant architectures and the inefficiencies of the existing architecture-agnostic methods. Building equivariant models using traditional methods requires designing equivariant versions of existing models and training them from scratch, a process that is both impractical and resource-intensive. Canonicalization has emerged as a promising alternative for inducing equivariance without altering model architecture, but it suffers from the need for highly expressive and expensive equivariant networks to learn canonical orientations accurately. We propose a new optimization-based method that employs any non-equivariant network for canonicalization. Our method uses contrastive learning to efficiently learn a canonical orientation and offers more flexibility for the choice of canonicalization network. We empirically demonstrate that this approach outperforms existing methods in achieving equivariance for large pretrained models and significantly speeds up the canonicalization process, making it up to 2 times faster.

\end{abstract}

%% file: sec/1_intro.tex
\section{Introduction}
\label{sec:intro}


Equivariant deep learning has emerged as a prominent approach within deep learning, aimed at developing neural networks that inherently understand and adapt to the symmetries in their input data \cite{lecun1995convolutional, cohen2016group, weiler2019general, cesa2021program, deng2021vector}. By constructing models that remain unaffected by transformations such as rotations or reflections, these networks preserve the core properties of the data, facilitating more efficient learning and better generalization across tasks. This notion of equivariance proves invaluable in areas such as computer vision \cite{yu2022rotationally,wu2023transformation,bekkers2018roto,chen2023imaging}, scientific applications \cite{kaba2022equivariant, duval2023faenet, bogatskiy2022symmetry, batatia2022mace, schutt2021equivariant}, graphs \cite{brandstetter2021geometric, duval2023hitchhiker, gasteiger2019directional, gasteiger2021gemnet}, and reinforcement learning \cite{mondal2020group, mondal2022eqr, van2020mdp, wang2022robot, wang2022equivariant, van2021multi}, where the ability to recognize patterns and make robust predictions demand a nuanced grasp of underlying data symmetries.

In the realm of equivariant model design, where the focus has traditionally been on creating novel equivariant layers \cite{cohen2016group, weiler2019general, cesa2021program, worrall2017harmonic,deng2021vector, cohen2018spherical}, a fresh research direction has emerged that centers around architecture-agnostic approaches. These methods, including symmetrization \cite{basu2023equi, basu2023efficient, kim2023probabilistic}, frame-averaging \cite{puny2021frame}, and canonicalization \cite{kaba2023equivariance, mondal2023equivariant}, aim to make models inherently equivariant to the transformation of the data without the need for specialized parameterized layers and activations. These methods significantly simplifies equivariant model design and, in some scenarios, make them more efficient. In particular, canonicalization proved to be a cheap and efficient way to any existing neural network equivariant to a group of transformations \cite{kaba2023equivariance}. This idea becomes more appealing especially when it comes to making any existing widely used large pre-trained models, including foundation models like SAM \cite{kirillov2023segment}, completely equivariant \cite{mondal2023equivariant}.

In this work, we focus on enhancing the canonicalization process, specifically addressing its fundamental limitation: the reliance on equivariant architecture for constructing the canonicalization network. We explore an alternate optimization approach and propose a novel method that uses contrastive learning during training to learn a unique canonical orientation for inference. Our technique gives us the flexibility to use any neural network as a canonicalization network, including pretrained ones that further improves the ease of optimization. This further relaxes any architectural constraints required to build equivariant models making them more accessible to the wider deep learning community. Moreover, we demonstrate that our simple approach not only outperforms existing method to build equivariant models using canonicalization but also makes canonicalization process significantly more effcient.




%% file: sec/2_background.tex
\section{Background}
\label{sec:background} 

\citet{kaba2023equivariance} introduces a systematic and general method for equivariant machine learning based on learning mappings to canonical samples. Rather than trying to hand-engineer these canonicalization functions, they propose to learn them in an end-to-end fashion with a prediction neural network. 
Canonicalization can be seamlessly integrated as an independent module into any existing architecture to make them equivariant to a wide range of transformation groups, discrete or continuous. This approach not only matches the expressive capabilities of methods like \textit{frame averaging} by \citet{puny2021frame} but also surpasses them by offering simplicity, efficiency, and a systematic end-to-end learning method that replaces hand-engineered frames with learned mappings for each group.

\begin{figure*}[t]
\centerline{\includegraphics[width=0.8\linewidth]{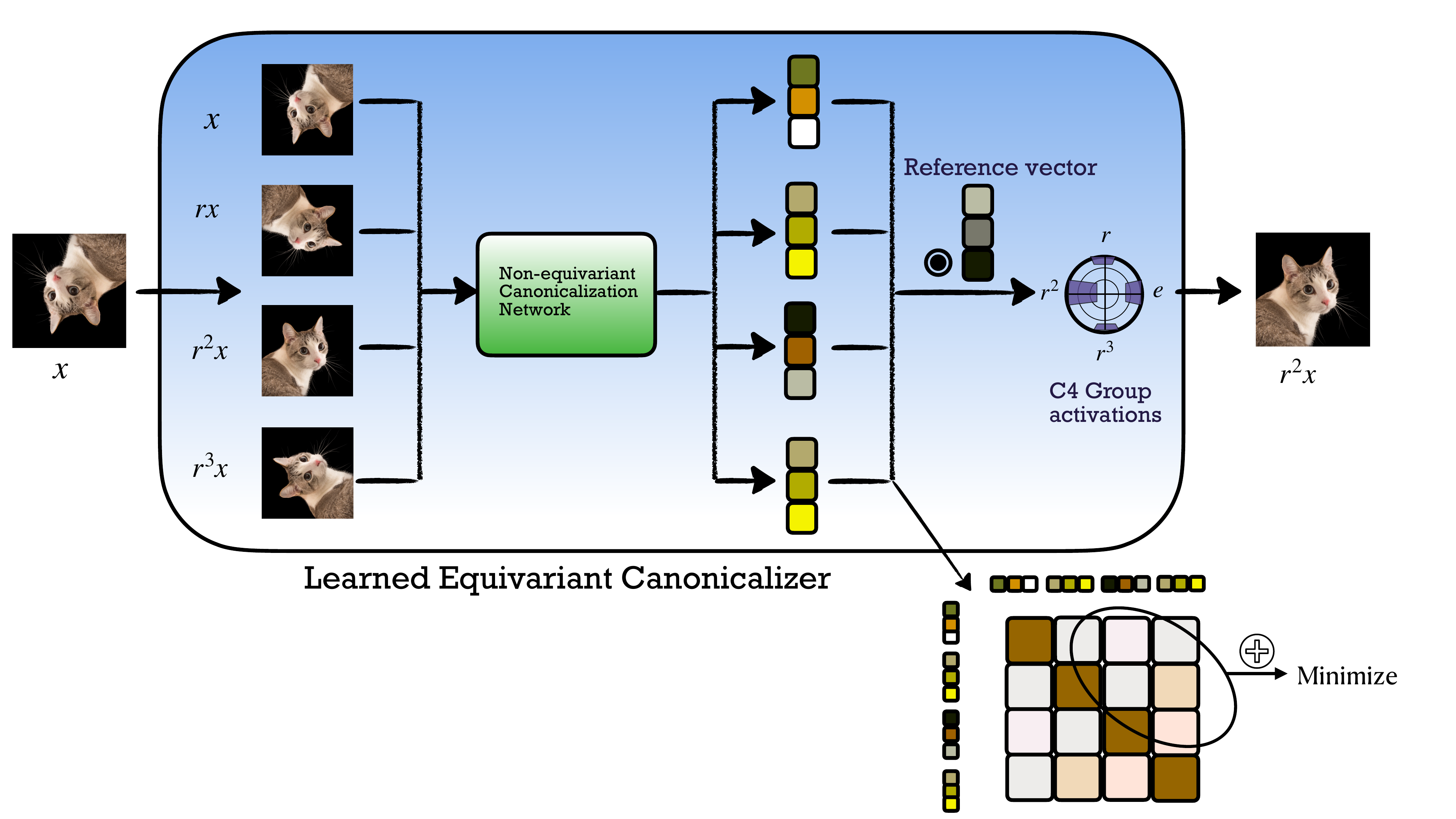}}
\caption{
Learning equivariant canonicalizer with a non-equivariant canonicalization network. All the transformations of the group are applied to the input image and passed through the canonicalization network in parallel. A dot product of the output of the canonicalization network with a reference vector gives us a distribution over the transformations to canonicalize the input. We also minimize the similarity between the vectors to get a unique canonical orientation.
}
\label{fig:method}
\end{figure*}

\subsection{Formulation}
The approach formulates the invariance requirement for a function as the capability to map all members of a group orbit to the same output. This is achieved by mapping inputs to a canonical sample from their orbit before applying the function. For equivariance, elements are also mapped to a canonical sample and, following function application, transformed back according to their original position in orbit. This can be formalized by writing the equivariant function $f$ in \textit{canonicalized form} as
\begin{align}
& f\pr{\v{x}} = c'\pr{\v{x}} \v{p}\pr{c\pr{\v{x}}^{-1} \v{x} } 
\label{eq:model}
\end{align}
where the function $\v{p}: \s{X}\to \s{Y}$ is called the \emph{prediction function} and the function $c: \s{X}\to \rho\pr{\g{G}}$ is called the \emph{canonicalization function}. Here $c\pr{\v{x}}^{-1}$ is the inverse of the representation matrix and $c'\pr{\v{x}} = \rho'\pr{\rho^{-1}\pr{c\pr{\v{x}}}}$ is the counterpart of $c\pr{\v{x}}$ on the output.

\citet{kaba2023equivariance} shows that $f$ is $\g{G}$-equivariant for any prediction function as long as the canonicalization function is itself $\g{G}$-equivariant, $c\pr{\rho\pr{g} \v{x}} = \rho\pr{g} c\pr{\v{x}} \ \forall \ g, x \in \g{G} \times \s{X}$. This effectively decouples the equivariance and prediction components. Moreover, they also introduce the concept of relaxed equivariance to deal with symmetric inputs in $\s{X}$.

\subsection{Canonicalization Function}
\citet{kaba2023equivariance} choose the canonicalization function to be any existing equivariant neural network architecture with the output being a group element, which they call the \textit{direct approach}. This ensures the $\g{G}$-equivariance constraint of the canonicalization function. For example, Group Convolutional Neural Network (G-CNNs) \cite{cohen2016group} are used to design a canonicalization function that is equivariant to the group of discrete rotations. 

They also provide an alternative \textit{optimization approach}, in which the canonicalization function is defined as
\begin{align}
\label{eq:optim}
c\pr{\v{x}} \in \argmin_{\rho\pr{g}\in \rho\pr{\g{G}}} s\pr{ \rho\pr{g}, \v{x}}
\end{align}
where $s : \rho\pr{\g{G}} \times \set{X}\to \mathbb{R}$ can be a neural network. In general, a set of elements can minimize $s$, from which one of them is chosen arbitrarily. The function $s$ has to satisfy the following equivariance condition
\begin{align}
s\pr{ \rho\pr{g}, \rho\pr{g_1}\v{x}} = s\pr{ \rho\pr{g_1}^{-1}\rho\pr{g}, \v{x}},  \forall g, g_1 \in \g{G}
\label{eq:optim_equiv}
\end{align}
and has to be such that argmin is a subset of a coset of the stabilizer of $\v{x}$. \footnote{minimum should be unique in each orbit up to some input symmetry} These are sufficient conditions for \cref{eq:optim} to be a suitable canonicalization function \cite{kaba2023equivariance}. The equivariance condition on $s$ can now not only be satisfied with equivariant architecture but also using a non-equivariant function $u: \s{X}\to \mathbb{R}$ and by defining:
\begin{align}
s\pr{ \rho\pr{g}, \v{x}} = u\pr{ \rho\pr{g}^{-1} \v{x}} \notag
\end{align}
In this paper, we use this to design a novel and simpler technique for learning an equivariant canonicalization function with any existing neural network.

\subsection{Prior Regularization}
\citet{mondal2023equivariant} extend canonicalization to adapt any existing pretrained neural network to its equivariant counterpart. To enhance the canonicalization process, ensuring input orientations closely match what's found in our training data, they introduce a novel regularizer known as the Canonicalization Prior (CP). This approach aims to leverage the similarity in orientations between fine-tuning and training datasets to guide canonicalization in closely matching the original orientations of inputs seen by the pretrained network during the pretraining stage.

From a probabilistic standpoint, the canonicalization function maps each data point into a probability distribution across a group of transformations, denoted by $\g{G}$. For a specific data point $\v{x}$, let $\mathbb{P}_{c(\v{x})}$ represent the distribution induced by the canonicalization function over $\g{G}$. Assuming a canonicalization prior exists for the dataset $\s{D}$, characterized by a distribution $\mathbb{P}_{\s{D}}$ over $\g{G}$, prior regularization aims to minimize the Kullback-Leibler (KL) divergence between $\mathbb{P}_{\s{D}}$ and $\mathbb{P}_{c(\v{x})}$. This leads to the loss function: $ \mathcal{L}_{\text{prior}} = \mathbb{E}_{\v{x}\sim\mathcal{D}} \left[D_{KL}(\mathbb{P}_{\s{D}} \parallel \mathbb{P}_{c(\v{x})})\right]$.

%% file: sec/3_method.tex
\section{Method}
\label{sec:method}

We extend the optimization approach to enable the use of any neural network for canonicalization, with a special focus on a group of discrete transformations in this work. The optimization formula for a discrete group, denoted by $\g{G}$, is:
\begin{align}
g \in \argmin_{g\in \g{G}} u\pr{ \rho\pr{g}^{-1} \v{x}}
\end{align}
Assuming there are no symmetric elements in the orbit represented by $\v{x}^{\g{G}} = \{ \rho\pr{g}^{-1} \v{x} \mid g\in \g{G}\}$, it is important to ensure the function $u()$ has a unique minimum to establish a canonical orientation. Additionally, should symmetric elements exist within the orbit, and if the minimum is attained among these symmetric positions, selecting any one of them will yield the correct canonical orientation (see \cite{kaba2023equivariance, kaba2023symmetry}).

In order to design this function $u()$, we resort to learning it using a neural network and minimizing the similarity among the output of the elements in the orbit. We output vectors corresponding to every element in the orbit using any neural network $s_{\theta}()$. This allows us to use techniques from the self-supervised learning literature to prevent representation collapse \cite{wang2020understanding, chen2020simple,balestriero2023cookbook} including non-contrastive ones that relies on the maximizing the eigenspectrum of the covariance matrix \cite{bardes2021vicreg, zbontar2021barlow}. In contrast to this, outputting scalars directly makes the optimization harder while limiting us to only contrastive methods. Then, we take a dot product of outputs of $s_{\theta}()$ with a reference vector $v_R$, which we can either learn or keep fixed. We get the distribution induced by canonicalization function $\mathbb{P}_{c(\v{x})}$ by taking a softmax over  $\{ v_R\cdot s_{\theta}\pr{\rho\pr{g}^{-1} \v{x}} / \tau \mid g\in \g{G} \}$, where $\tau$ is the temperature parameter of the distribution that controls its sharpness and is set to $1$ in our experiments. In this formulation, $u()$ becomes the probability mass function. The final optimization formulation becomes:

\begin{align}
g \in \argmin_{g\in \g{G}} \frac{\exp{(v_R\cdot s_{\theta}\pr{ \rho\pr{g}^{-1} \v{x}} / \tau)}}{ \sum_{g'\in \g{G}} \exp{\pr{v_R\cdot s_{\theta}\pr{ \rho\pr{g'}^{-1} \v{x}} / \tau}}} 
\end{align}

Inorder to make this canonicalization process differentiable, we use straight through gradient trick as proposed in \cite{kaba2023equivariance}. Alternatively, to introduce more augmentation effect during training \cite{mondal2023equivariant}, one can use Gumbel Softmax \cite{jang2016categorical} to sample from $\mathbb{P}_{c(\v{x})}$ in a differentiable way. Now, to obtain an unique canonical orientation, we train $s_{\theta}()$ to output different vectors for every unique element in the orbit $\v{x}^{\g{G}}$ by minimizing the following loss, $\mathcal{L}_{Opt} $: 

\begin{align}
 \EE_{\v{x} \in \mathcal{D}}\br{ \sum_{g_i, g_j\in \g{G}, g_i\neq g_j} s_{\theta}\pr{ \rho\pr{g_i}^{-1} \v{x}} \cdot s_{\theta}\pr{ \rho\pr{g_j}^{-1} \v{x}}}
\label{eqn:opt loss}
\end{align}

where $\mathcal{D}$ is the training dataset. This loss prevents the collapse of learnt vectors in the output space of $s_{\theta}()$ for different transformations of the input $x$ by minimizing their similarity measured using elementwise dot product. \cref{fig:method} shows a schematic of our simple approach. The use of non-contrastive approaches \cite{zbontar2021barlow, bardes2021vicreg} that uses the cross-correlation between these vectors to prevent representation collapse is an interesting avenue of future work.

In the context of training from scratch \cite{kaba2023equivariance}, the loss from \cref{eqn:opt loss} can be jointly optimized with the task loss. Similarly, for fine-tuning or zero-shot adaptation \cite{mondal2023equivariant}, an additional prior regularization loss is used.
Assuming the identity transformation to be the prior for natural image dataset \cite{mondal2023equivariant}, the loss $\mathcal{L}_{prior}$ is given by:

\begin{align}
\EE_{\v{x} \in \mathcal{D}_f}\br{ - \log \pr{ \frac{\exp{(v_R\cdot s_{\theta}\pr{ \v{x}}/ \tau})}{ \sum_{g\in \g{G}} \exp{\pr{v_R\cdot s_{\theta}\pr{ \rho\pr{g}^{-1} \v{x}}/ \tau}}} }}
\end{align}

where $\mathcal{D}_f$ is the finetuning dataset. As this formulation transfers the equivariance constraint of \cref{eq:optim_equiv} to minimizing the loss in \cref{eqn:opt loss} over the data distribution, we can conveniently start with a pretrained $s_{\theta}()$ to further ease the optimization process.   

Typically, we choose $s_{\theta}()$ that are smaller and faster than the large prediction network $\v{p}()$. This is based on the assumption that determining a canonical orientation is simpler than the more complex downstream task that demands a deeper understanding of the input. Therefore, our method requires $|\g{G}|$ forward passes in parallel through $s_{\theta}()$ instead of the prediction function $\v{p}()$, making it significantly more efficient than symmetrization-based methods \cite{basu2023equi,basu2023efficient,puny2021frame}. 

%% file: sec/4_results.tex
\section{Results}
\begin{table*}[t]
\centering
\begin{tabular}{cccccc}
    \toprule
    \midrule
    \multicolumn{2}{c}{Pretrained Large Prediction Network $\to$} & \multicolumn{2}{c}{ResNet50} & \multicolumn{2}{c}{ViT} \\
    \midrule
    Datasets $\downarrow$ & Model & Acc & $C4$-Avg Acc & Acc & $C4$-Avg Acc \\
    \midrule
    
   \multirow{ 4}{*}{CIFAR10} 
    & Vanilla & \textbf{97.33 $\pm$ 0.01} & 69.72 $\pm$ 0.25 & \textbf{98.13 $\pm$ 0.04} & 68.98 $\pm$ 0.48\\
    & $C4$-Augmentation & 95.76 $\pm$ 0.01 & 94.77 $\pm$ 0.05 & 96.61 $\pm$ 0.04 & 95.60 $\pm$ 0.03 \\
     & \makecell{EquiAdapt} & 96.19 $\pm$ 0.01 & 96.18 $\pm$ 0.02 & 96.14 $\pm$ 0.14 & 96.12 $\pm$ 0.11\\
     & \makecell{EquiOptAdapt} 
     & 97.16 $\pm$ 0.01 & \textbf{97.16 $\pm$ 0.01} & 96.96 $\pm$ 0.02 & \textbf{96.96 $\pm$ 0.02}
     \\

\midrule
     
   \multirow{ 4}{*}{STL10} 
    & Vanilla & \textbf{98.30 $\pm$ 0.01} & 88.61 $\pm$ 0.34 & \textbf{98.31 $\pm$ 0.09} & 78.63 $\pm$ 0.25 \\
    & $C4$-Augmentation & 98.20 $\pm$ 0.05 & 95.84 $\pm$ 0.04 & 97.69 $\pm$ 0.07 & 95.79 $\pm$ 0.14 \\
    & \makecell{EquiAdapt} 
        & 97.01 $\pm$ 0.01 & 96.98 $\pm$ 0.02
        & 96.15 $\pm$ 0.05 & 96.15 $\pm$ 0.05 \\
    & \makecell{EquiOptAdapt} 
        & 98.04 $\pm$ 0.05 & \textbf{98.04 $\pm$ 0.04}
        & 97.32 $\pm$ 0.01 & \textbf{97.32 $\pm$ 0.01} \\

    \bottomrule
  \end{tabular}
  \caption{Performance comparison of large pretrained models finetuned on different vision datasets. Both Accuracy (Acc) and $C4$-Average Accuracy ($C4$-Avg Acc) are reported. Acc refers to the accuracy on the original test set, and $C4$-Avg Acc refers to the accuracy on the augmented test set obtained using the group $C_4$.}
  \label{tab:classification}
\end{table*}

While our method applies to training any equivariant models from scratch, motivated by the practical advantages of using large scale pretrained models, we only focus on their equivariant adaptation by finetuning them using prior regularization loss. This section presents results from experiments on well-known, publicly available pretrained networks. Our method, EquiOptAdapt, enables equivariant adaptation of these models without any additional architecture constraints on the canonicalizer. EquiOptAdapt maintains fine-tuned model performance, increases robustness against known out-of-distribution transformations, and operates faster than conventional equivariant canonicalization approaches.

\subsection{Image Classification}
\label{subsec:image-classification}

\begin{table*}[t]
\small
\centering
\vspace{1em}
\resizebox{\textwidth}{!}{
\begin{tabular}{c|cc|cc|cc}
\toprule

\textbf{Network ($\to$)} & \multicolumn{2}{c|}{\textbf{MaskRCNN}} & \multicolumn{2}{c|}{\textbf{SAM}} & \textbf{MaskRCNN} & \textbf{SAM} \\

\cmidrule{2-3} \cmidrule{4-5} \cmidrule{6-7}
\textbf{Setup ($\downarrow$)} & mAP & C4-Avg mAP & mAP & C4-Avg mAP & \multicolumn{2}{c}{Inference times ($\downarrow$)} \\

\midrule
Zero-shot & \textbf{48.19} & 29.34 & \textbf{62.32} & 58.77 & 23m 53s & 2h 28m 43s \\
EquiAdapt & 46.80 & 46.79 & 62.10 & 62.10 & 27m 09s (\textcolor{red}{+13.68\%}) & 2h 34m 36s (\textcolor{red}{+3.96\%}) \\
EquiOptAdapt & 48.01 & \textbf{48.01} & 62.30 & \textbf{62.30} & 25m 35s (\textcolor{subduedgreen}{+7.12\%}) & 2h 30m 42s (\textcolor{subduedgreen}{+1.33\%}) \\
\bottomrule

\end{tabular}
}
\caption{Zero-shot performance comparison and inference times of large pretrained segmentation models with and without trained canonicalization functions on the validation set of COCO 2017 dataset \cite{lin2014microsoft}.}
\label{tab:instance_segmentation_combined}

\end{table*}

\paragraph{Experiment Setup.} The Vanilla setup consists of fine-tuning ResNet50 \cite{he2016deep} and Vision Transformer (ViT, \cite{dosovitskiy2020image}), which are widely used for obtaining image embeddings to solve downstream tasks. Both architectures were pretrained on ImageNet-1K \cite{deng2009imagenet}, and the checkpoints are publicly available. \footnote{\href{https://pytorch.org/vision/stable/models/generated/torchvision.models.resnet50.html}{Resnet50 checkpoint from PyTorch}} \footnote{\href{https://pytorch.org/vision/stable/models/generated/torchvision.models.vit\_b\_16.html}{ViT-B/16 checkpoint from PyTorch}}. Another strong baseline is to fine-tune the pretrained architecture using $C_4$ group data augmentation, given our prior knowledge that the evaluation is performed on a $C4$-augmented test set. 


The \text{EquiAdapt} setup \cite{mondal2023equivariant} uses an equivariant canonicalization network to build a canonicalizer that is placed before the pretrained architecture. Both the networks are finetuned using a cross-entropy loss for the classification task and an additional prior regularization loss is used for the canonicalization network. In comparison to this, the canonicalizer in EquiOptAdapt uses a smaller pretrained ResNet architecture as a canonicalization network $s_{\theta}()$. We set the output space of $s_{\theta}()$ to 128 dimension, and $v_{R}$ is a random constant Gaussian vector of the same dimension. Along with the cross-entropy classification task loss and $\mathcal{L}_{prior}$, the final fine-tuning loss includes $\mathcal{L}_{opt}$ to learn an equivariant canonicalizer.


\paragraph{Evaluation setup.} We use a similar evaluation protocol as \citet{mondal2023equivariant}. Along with the accuracy on the original test set, we use $C4$-Average Accuracy that indicates accuracy on an augmented test set, where each image in the test set was rotated with elements of $C_4$ group, i.e., group of 4 discrete rotations. 

\paragraph{Results.} We present the finetuning results for different setups in \cref{tab:classification} for CIFAR10 \cite{krizhevsky2009learning} and STL10 \cite{coates2011analysis}. Our findings demonstrate that both EquiOptAdapt and EquiAdapt exhibit comparable performance to the Vanilla setup in terms of test-set accuracy, with EquiOptAdapt showcasing superior performance. This suggests that pretrained non-equivariant canonicalization network can further ease the optimization, thereby enhancing their ability to learn the mapping from data input to a unique element within the orbit of the considered group. Similar to \citet{mondal2023equivariant}, we observe that more expressive canonicalizers lead to higher performance. Further, there is no gap between accuracy and $C4$-average accuracy, demonstrating the successful learning of equivariant canonicalizer, and hence, equivariant adaptation of the considered models. The Vanilla and C-4 Augmentation models perform significantly worse than equivariant adaptation based models while testing on C-4 augmented test set.

\subsection{Zero-shot Instance Segmentation}
\label{subsec:instance-segmentation}
\paragraph{Experiment Setup.} Next, we compare the zero-shot instance segmentation results for MaskRCNN \cite{he2017mask} and Segment-Anything Model (SAM, \cite{kirillov2023segment}) on COCO 2017 \cite{lin2014microsoft}. Particularly, we evaluate promptable instance segmentation for the SAM, with bounding boxes as prompts. We keep the same setups as \cref{subsec:image-classification} where fine-tuning is replaced with zero-shot performance. Similar to the strategy in \citet{mondal2023equivariant}, where a canonicalizer is trained on the COCO dataset with prior regularization $\mathcal{L}_{prior}$, we only train our canonicalizer with an additional optimaztion loss $\mathcal{L}_{opt}$ to make the canonicalization process equivariant.
Similar to \cref{subsec:image-classification}, we initialize our non-equivariant canonicalizers with pretrained WideResNet-50 architecture.

\paragraph{Evaluation setup.} We use the mean-average precision (mAP) and $C4$-Average mAP scores. Here, again, $C4$-Average mAP score indicates the mAP score on an augmented \textit{val} set of COCO 2017, where each image (and bounding boxes) was rotated with elements of $C_4$ group while mAP indicates the mAP score on the original \textit{val} set. 

We also compare the relative wall clock time (in minutes) to learn the prior distribution $\mathbb{P}_{c(x)}$ during training with \cite{mondal2023equivariant}. Given that our chosen $\mathbb{P}_{c(x)}$ is effectively a $\delta$-distribution centred on the identity element $e$ of the group, we evaluate the accuracy of learning this prior as the identity metric.

\paragraph{Results.}
The results for various setups are presented in \Cref{tab:instance_segmentation_combined}. Our analysis reveals that EquiAdapt and EquiOptAdapt effectively achieve architecture-agnostic equivariant adaptation of large pretrained models while maintaining their mean Average Precision (mAP) performance. Notably, again, EquiOptAdapt outperforms EquiAdapt in this regard. Additionally, we provide comprehensive insights into the total inference times for each setup in \cref{tab:instance_segmentation_combined}. The inference times for EquiOptAdapt and EquiAdapt indicate that the canonicalization process is 2$\times$ faster for EquiOptAdapt. 

\begin{figure}[h]
\centerline{\includegraphics[width=\columnwidth]{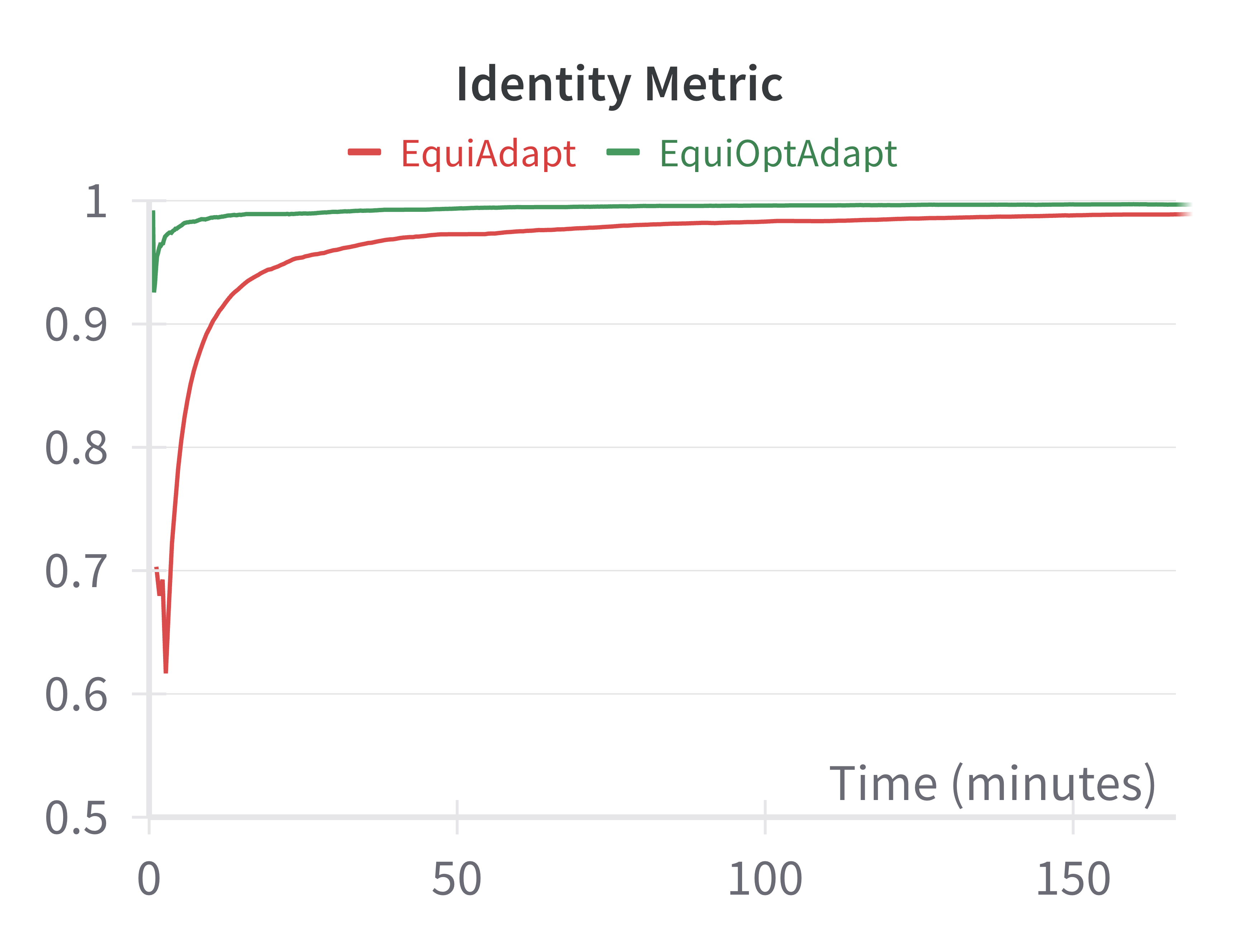}}
\caption{
Identity metric vs. Relative wall-time (in minutes). We define the identity metric as the percentage of input images mapped to the identity group element $e$, which is our prior distribution $\mathbb{P}_{c(x)}$. This figure demonstrates that our EquiOptAdapt is able to learn the prior faster than EquiAdapt.
}
\label{fig:identity-metric}
\end{figure}

Moreover, \Cref{fig:identity-metric} plots the relative wall-time for EquiOptAdapt and EquiAdapt against the identity metric. We demonstrate that our proposed EquiOptAdapt is able to learn the prior distribution faster than EquiAdapt. This results from the ability to use any exisiting non-equivariant pretrained WideResNet model that trains and run faster than an Equivariant WideResNet architecture used in EquiAdapt \cite{mondal2023equivariant}.  Therefore, our findings suggest that EquiOptAdapt generally offers better performance and faster training and inference times compared to EquiAdapt.

%% file: sec/5_discussion.tex
\section{Conclusion}

Generalizing to out-of-distribution data remains a considerable obstacle for state-of-the-art deep learning models, particularly due to input transformations like rotations, scalings, and orientation changes. Large pretrained models can be made equivariant to such transformations through canonicalization \cite{mondal2023equivariant}. However, existing approaches such as \cite{kaba2023equivariance,mondal2023equivariant} use equivariant networks for canonicalization which acts as a bottleneck for learning canonical orientations. This paper proposes EquiOptAdapt to address this expressivity constraint by leveraging an optimization-based approach with contrastive learning techniques enabling the use of any neural network architecture for canonicalization. Our experiments show that EquiOptAdapt preserves the performance of large pretrained models and surpasses existing methods on robust generalization to transformations of the data while significantly accelerating the canonicalization process. These findings highlight the practicality and effectiveness of our approach in achieving robust equivariant adaptation, marking an important advancement in improving out-of-distribution generalization and equivariant model design.

%% file: sec/6_limitations_future_work.tex
\section{Limitations and Future Work}

An important limitation of our current work lies in its focus on the group of discrete transformations. Prior experiments with continuous groups, such as the group of 2D rotations  $SO(2)$ \cite{mondal2023equivariant}, have revealed the limited ability of $E(2)$ steerable networks \cite{weiler2019general} to learn mappings from inputs to canonical orientations with prior regularization. This limitation can be potentially mitigated by utilizing more expressive unconstrained pretrained neural networks as the canonicalization network, which could lead to enhanced optimization. However, using continuous group will require test time optimization using the output energy values, which can make inference significantly more expensive. We plan to investigate this to find a workaround and introduce continuous rotations in future work. 

In addition to continuous rotations, we intend to incorporate higher-order discrete rotations and compare them. The finer rotation angles present an intriguing challenge for both continuous and higher-order discrete rotations due to the artifacts introduced at the corners of images. To address this, we aim to design novel techniques to make the canonicalization network robust to the effect of artifacts. Moreover, exploring other non-contrastive correlation-based methods to train the canonicalizer is another interesting direction for future research. 

Finally, automating prior discovery based on the performance of the pretrained model over different transformations of the input in the fine-tuning data can significantly impact the current limitation of manually deciding the prior. This can make the Equivariant Adaptation technique more general and agnostic to the choice of model, task, and data. 

%% file: main.bbl
\begin{thebibliography}{47}
\providecommand{\natexlab}[1]{#1}
\providecommand{\url}[1]{\texttt{#1}}
\expandafter\ifx\csname urlstyle\endcsname\relax
  \providecommand{\doi}[1]{doi: #1}\else
  \providecommand{\doi}{doi: \begingroup \urlstyle{rm}\Url}\fi

\bibitem[Balestriero et~al.(2023)Balestriero, Ibrahim, Sobal, Morcos, Shekhar, Goldstein, Bordes, Bardes, Mialon, Tian, et~al.]{balestriero2023cookbook}
Randall Balestriero, Mark Ibrahim, Vlad Sobal, Ari Morcos, Shashank Shekhar, Tom Goldstein, Florian Bordes, Adrien Bardes, Gregoire Mialon, Yuandong Tian, et~al.
\newblock A cookbook of self-supervised learning.
\newblock \emph{arXiv preprint arXiv:2304.12210}, 2023.

\bibitem[Bardes et~al.(2021)Bardes, Ponce, and LeCun]{bardes2021vicreg}
Adrien Bardes, Jean Ponce, and Yann LeCun.
\newblock Vicreg: Variance-invariance-covariance regularization for self-supervised learning.
\newblock \emph{arXiv preprint arXiv:2105.04906}, 2021.

\bibitem[Basu et~al.(2023{\natexlab{a}})Basu, Katdare, Sattigeri, Chenthamarakshan, Driggs-Campbell, Das, and Varshney]{basu2023efficient}
Sourya Basu, Pulkit Katdare, Prasanna Sattigeri, Vijil Chenthamarakshan, Katherine Driggs-Campbell, Payel Das, and Lav~R Varshney.
\newblock Efficient equivariant transfer learning from pretrained models.
\newblock In \emph{Advances in Neural Information Processing Systems}, pages 4213--4224. Curran Associates, Inc., 2023{\natexlab{a}}.

\bibitem[Basu et~al.(2023{\natexlab{b}})Basu, Sattigeri, Ramamurthy, Chenthamarakshan, Varshney, Varshney, and Das]{basu2023equi}
Sourya Basu, Prasanna Sattigeri, Karthikeyan~Natesan Ramamurthy, Vijil Chenthamarakshan, Kush~R Varshney, Lav~R Varshney, and Payel Das.
\newblock Equi-tuning: Group equivariant fine-tuning of pretrained models.
\newblock In \emph{Proceedings of the AAAI Conference on Artificial Intelligence}, pages 6788--6796, 2023{\natexlab{b}}.

\bibitem[Batatia et~al.(2022)Batatia, Kovacs, Simm, Ortner, and Cs{\'a}nyi]{batatia2022mace}
Ilyes Batatia, David~P Kovacs, Gregor Simm, Christoph Ortner, and G{\'a}bor Cs{\'a}nyi.
\newblock Mace: Higher order equivariant message passing neural networks for fast and accurate force fields.
\newblock \emph{Advances in Neural Information Processing Systems}, 35:\penalty0 11423--11436, 2022.

\bibitem[Bekkers et~al.(2018)Bekkers, Lafarge, Veta, Eppenhof, Pluim, and Duits]{bekkers2018roto}
Erik~J Bekkers, Maxime~W Lafarge, Mitko Veta, Koen~AJ Eppenhof, Josien~PW Pluim, and Remco Duits.
\newblock Roto-translation covariant convolutional networks for medical image analysis.
\newblock In \emph{Medical Image Computing and Computer Assisted Intervention--MICCAI 2018: 21st International Conference, Granada, Spain, September 16-20, 2018, Proceedings, Part I}, pages 440--448. Springer, 2018.

\bibitem[Bogatskiy et~al.(2022)Bogatskiy, Ganguly, Kipf, Kondor, Miller, Murnane, Offermann, Pettee, Shanahan, Shimmin, et~al.]{bogatskiy2022symmetry}
Alexander Bogatskiy, Sanmay Ganguly, Thomas Kipf, Risi Kondor, David~W Miller, Daniel Murnane, Jan~T Offermann, Mariel Pettee, Phiala Shanahan, Chase Shimmin, et~al.
\newblock Symmetry group equivariant architectures for physics.
\newblock \emph{arXiv preprint arXiv:2203.06153}, 2022.

\bibitem[Brandstetter et~al.(2021)Brandstetter, Hesselink, van~der Pol, Bekkers, and Welling]{brandstetter2021geometric}
Johannes Brandstetter, Rob Hesselink, Elise van~der Pol, Erik~J Bekkers, and Max Welling.
\newblock Geometric and physical quantities improve e (3) equivariant message passing.
\newblock In \emph{International Conference on Learning Representations}, 2021.

\bibitem[Cesa et~al.(2021)Cesa, Lang, and Weiler]{cesa2021program}
Gabriele Cesa, Leon Lang, and Maurice Weiler.
\newblock A program to build e (n)-equivariant steerable cnns.
\newblock In \emph{International conference on learning representations}, 2021.

\bibitem[Chen et~al.(2023)Chen, Davies, Ehrhardt, Sch{\"o}nlieb, Sherry, and Tachella]{chen2023imaging}
Dongdong Chen, Mike Davies, Matthias~J Ehrhardt, Carola-Bibiane Sch{\"o}nlieb, Ferdia Sherry, and Juli{\'a}n Tachella.
\newblock Imaging with equivariant deep learning: From unrolled network design to fully unsupervised learning.
\newblock \emph{IEEE Signal Processing Magazine}, 40\penalty0 (1):\penalty0 134--147, 2023.

\bibitem[Chen et~al.(2020)Chen, Kornblith, Norouzi, and Hinton]{chen2020simple}
Ting Chen, Simon Kornblith, Mohammad Norouzi, and Geoffrey Hinton.
\newblock A simple framework for contrastive learning of visual representations.
\newblock In \emph{International conference on machine learning}, pages 1597--1607. PMLR, 2020.

\bibitem[Coates et~al.(2011)Coates, Ng, and Lee]{coates2011analysis}
Adam Coates, Andrew Ng, and Honglak Lee.
\newblock An analysis of single-layer networks in unsupervised feature learning.
\newblock In \emph{Proceedings of the fourteenth international conference on artificial intelligence and statistics}, pages 215--223. JMLR Workshop and Conference Proceedings, 2011.

\bibitem[Cohen and Welling(2016)]{cohen2016group}
Taco Cohen and Max Welling.
\newblock Group equivariant convolutional networks.
\newblock In \emph{International conference on machine learning}, pages 2990--2999. PMLR, 2016.

\bibitem[Cohen et~al.(2018)Cohen, Geiger, K{\"o}hler, and Welling]{cohen2018spherical}
Taco~S Cohen, Mario Geiger, Jonas K{\"o}hler, and Max Welling.
\newblock Spherical cnns.
\newblock \emph{arXiv preprint arXiv:1801.10130}, 2018.

\bibitem[Deng et~al.(2021)Deng, Litany, Duan, Poulenard, Tagliasacchi, and Guibas]{deng2021vector}
Congyue Deng, Or Litany, Yueqi Duan, Adrien Poulenard, Andrea Tagliasacchi, and Leonidas~J Guibas.
\newblock Vector neurons: A general framework for so (3)-equivariant networks.
\newblock In \emph{Proceedings of the IEEE/CVF International Conference on Computer Vision}, pages 12200--12209, 2021.

\bibitem[Deng et~al.(2009)Deng, Dong, Socher, Li, Li, and Fei-Fei]{deng2009imagenet}
Jia Deng, Wei Dong, Richard Socher, Li-Jia Li, Kai Li, and Li Fei-Fei.
\newblock Imagenet: A large-scale hierarchical image database.
\newblock In \emph{2009 IEEE conference on computer vision and pattern recognition}, pages 248--255. Ieee, 2009.

\bibitem[Dosovitskiy et~al.(2020)Dosovitskiy, Beyer, Kolesnikov, Weissenborn, Zhai, Unterthiner, Dehghani, Minderer, Heigold, Gelly, et~al.]{dosovitskiy2020image}
Alexey Dosovitskiy, Lucas Beyer, Alexander Kolesnikov, Dirk Weissenborn, Xiaohua Zhai, Thomas Unterthiner, Mostafa Dehghani, Matthias Minderer, Georg Heigold, Sylvain Gelly, et~al.
\newblock An image is worth 16x16 words: Transformers for image recognition at scale.
\newblock In \emph{International Conference on Learning Representations}, 2020.

\bibitem[Duval et~al.(2023{\natexlab{a}})Duval, Mathis, Joshi, Schmidt, Miret, Malliaros, Cohen, Li{\`o}, Bengio, and Bronstein]{duval2023hitchhiker}
Alexandre Duval, Simon~V Mathis, Chaitanya~K Joshi, Victor Schmidt, Santiago Miret, Fragkiskos~D Malliaros, Taco Cohen, Pietro Li{\`o}, Yoshua Bengio, and Michael Bronstein.
\newblock A hitchhiker's guide to geometric gnns for 3d atomic systems.
\newblock \emph{arXiv preprint arXiv:2312.07511}, 2023{\natexlab{a}}.

\bibitem[Duval et~al.(2023{\natexlab{b}})Duval, Schmidt, Hern{\'a}ndez-Garćia, Miret, Malliaros, Bengio, and Rolnick]{duval2023faenet}
Alexandre~Agm Duval, Victor Schmidt, Alex Hern{\'a}ndez-Garćia, Santiago Miret, Fragkiskos~D Malliaros, Yoshua Bengio, and David Rolnick.
\newblock Faenet: Frame averaging equivariant gnn for materials modeling.
\newblock In \emph{International Conference on Machine Learning}, pages 9013--9033. PMLR, 2023{\natexlab{b}}.

\bibitem[Gasteiger et~al.(2019)Gasteiger, Gro{\ss}, and G{\"u}nnemann]{gasteiger2019directional}
Johannes Gasteiger, Janek Gro{\ss}, and Stephan G{\"u}nnemann.
\newblock Directional message passing for molecular graphs.
\newblock In \emph{International Conference on Learning Representations}, 2019.

\bibitem[Gasteiger et~al.(2021)Gasteiger, Becker, and G{\"u}nnemann]{gasteiger2021gemnet}
Johannes Gasteiger, Florian Becker, and Stephan G{\"u}nnemann.
\newblock Gemnet: Universal directional graph neural networks for molecules.
\newblock \emph{Advances in Neural Information Processing Systems}, 34:\penalty0 6790--6802, 2021.

\bibitem[He et~al.(2016)He, Zhang, Ren, and Sun]{he2016deep}
Kaiming He, Xiangyu Zhang, Shaoqing Ren, and Jian Sun.
\newblock Deep residual learning for image recognition.
\newblock In \emph{Proceedings of the IEEE conference on computer vision and pattern recognition}, pages 770--778, 2016.

\bibitem[He et~al.(2017)He, Gkioxari, Doll{\'a}r, and Girshick]{he2017mask}
Kaiming He, Georgia Gkioxari, Piotr Doll{\'a}r, and Ross Girshick.
\newblock Mask r-cnn.
\newblock In \emph{Proceedings of the IEEE international conference on computer vision}, pages 2961--2969, 2017.

\bibitem[Jang et~al.(2016)Jang, Gu, and Poole]{jang2016categorical}
Eric Jang, Shixiang Gu, and Ben Poole.
\newblock Categorical reparameterization with gumbel-softmax.
\newblock \emph{arXiv preprint arXiv:1611.01144}, 2016.

\bibitem[Kaba and Ravanbakhsh(2022)]{kaba2022equivariant}
Oumar Kaba and Siamak Ravanbakhsh.
\newblock Equivariant networks for crystal structures.
\newblock \emph{Advances in Neural Information Processing Systems}, 35:\penalty0 4150--4164, 2022.

\bibitem[Kaba and Ravanbakhsh(2023)]{kaba2023symmetry}
S{\'e}kou-Oumar Kaba and Siamak Ravanbakhsh.
\newblock Symmetry breaking and equivariant neural networks.
\newblock \emph{arXiv preprint arXiv:2312.09016}, 2023.

\bibitem[Kaba et~al.(2023)Kaba, Mondal, Zhang, Bengio, and Ravanbakhsh]{kaba2023equivariance}
S{\'e}kou-Oumar Kaba, Arnab~Kumar Mondal, Yan Zhang, Yoshua Bengio, and Siamak Ravanbakhsh.
\newblock Equivariance with learned canonicalization functions.
\newblock In \emph{International Conference on Machine Learning}, pages 15546--15566. PMLR, 2023.

\bibitem[Kim et~al.(2023)Kim, Nguyen, Suleymanzade, An, and Hong]{kim2023probabilistic}
Jinwoo Kim, Dat Nguyen, Ayhan Suleymanzade, Hyeokjun An, and Seunghoon Hong.
\newblock Learning probabilistic symmetrization for architecture agnostic equivariance.
\newblock In \emph{Advances in Neural Information Processing Systems}, pages 18582--18612. Curran Associates, Inc., 2023.

\bibitem[Kirillov et~al.(2023)Kirillov, Mintun, Ravi, Mao, Rolland, Gustafson, Xiao, Whitehead, Berg, Lo, et~al.]{kirillov2023segment}
Alexander Kirillov, Eric Mintun, Nikhila Ravi, Hanzi Mao, Chloe Rolland, Laura Gustafson, Tete Xiao, Spencer Whitehead, Alexander~C Berg, Wan-Yen Lo, et~al.
\newblock Segment anything.
\newblock In \emph{Proceedings of the IEEE/CVF International Conference on Computer Vision}, pages 4015--4026, 2023.

\bibitem[Krizhevsky et~al.(2009)]{krizhevsky2009learning}
Alex Krizhevsky et~al.
\newblock Learning multiple layers of features from tiny images.
\newblock 2009.

\bibitem[LeCun et~al.()LeCun, Bengio, et~al.]{lecun1995convolutional}
Yann LeCun, Yoshua Bengio, et~al.
\newblock Convolutional networks for images, speech, and time series.

\bibitem[Lin et~al.(2014)Lin, Maire, Belongie, Hays, Perona, Ramanan, Doll{\'a}r, and Zitnick]{lin2014microsoft}
Tsung-Yi Lin, Michael Maire, Serge Belongie, James Hays, Pietro Perona, Deva Ramanan, Piotr Doll{\'a}r, and C~Lawrence Zitnick.
\newblock Microsoft coco: Common objects in context.
\newblock In \emph{Computer Vision--ECCV 2014: 13th European Conference, Zurich, Switzerland, September 6-12, 2014, Proceedings, Part V 13}, pages 740--755. Springer, 2014.

\bibitem[Mondal et~al.(2020)Mondal, Nair, and Siddiqi]{mondal2020group}
Arnab~Kumar Mondal, Pratheeksha Nair, and Kaleem Siddiqi.
\newblock Group equivariant deep reinforcement learning.
\newblock \emph{arXiv preprint arXiv:2007.03437}, 2020.

\bibitem[Mondal et~al.(2022)Mondal, Jain, Siddiqi, and Ravanbakhsh]{mondal2022eqr}
Arnab~Kumar Mondal, Vineet Jain, Kaleem Siddiqi, and Siamak Ravanbakhsh.
\newblock Eqr: Equivariant representations for data-efficient reinforcement learning.
\newblock In \emph{International Conference on Machine Learning}, pages 15908--15926. PMLR, 2022.

\bibitem[Mondal et~al.(2023)Mondal, Panigrahi, Kaba, Mudumba, and Ravanbakhsh]{mondal2023equivariant}
Arnab~Kumar Mondal, Siba~Smarak Panigrahi, Oumar Kaba, Sai~Rajeswar Mudumba, and Siamak Ravanbakhsh.
\newblock Equivariant adaptation of large pretrained models.
\newblock In \emph{Advances in Neural Information Processing Systems}, pages 50293--50309. Curran Associates, Inc., 2023.

\bibitem[Puny et~al.(2021)Puny, Atzmon, Ben-Hamu, Misra, Grover, Smith, and Lipman]{puny2021frame}
Omri Puny, Matan Atzmon, Heli Ben-Hamu, Ishan Misra, Aditya Grover, Edward~J Smith, and Yaron Lipman.
\newblock Frame averaging for invariant and equivariant network design.
\newblock \emph{arXiv preprint arXiv:2110.03336}, 2021.

\bibitem[Sch{\"u}tt et~al.(2021)Sch{\"u}tt, Unke, and Gastegger]{schutt2021equivariant}
Kristof Sch{\"u}tt, Oliver Unke, and Michael Gastegger.
\newblock Equivariant message passing for the prediction of tensorial properties and molecular spectra.
\newblock In \emph{International Conference on Machine Learning}, pages 9377--9388. PMLR, 2021.

\bibitem[Van~der Pol et~al.(2020)Van~der Pol, Worrall, van Hoof, Oliehoek, and Welling]{van2020mdp}
Elise Van~der Pol, Daniel Worrall, Herke van Hoof, Frans Oliehoek, and Max Welling.
\newblock Mdp homomorphic networks: Group symmetries in reinforcement learning.
\newblock \emph{Advances in Neural Information Processing Systems}, 33:\penalty0 4199--4210, 2020.

\bibitem[van~der Pol et~al.(2021)van~der Pol, van Hoof, Oliehoek, and Welling]{van2021multi}
Elise van~der Pol, Herke van Hoof, Frans~A Oliehoek, and Max Welling.
\newblock Multi-agent mdp homomorphic networks.
\newblock \emph{arXiv preprint arXiv:2110.04495}, 2021.

\bibitem[Wang et~al.(2022{\natexlab{a}})Wang, Jia, Zhu, Walters, and Platt]{wang2022robot}
Dian Wang, Mingxi Jia, Xupeng Zhu, Robin Walters, and Robert Platt.
\newblock On-robot learning with equivariant models.
\newblock \emph{arXiv preprint arXiv:2203.04923}, 2022{\natexlab{a}}.

\bibitem[Wang et~al.(2022{\natexlab{b}})Wang, Walters, Zhu, and Platt]{wang2022equivariant}
Dian Wang, Robin Walters, Xupeng Zhu, and Robert Platt.
\newblock Equivariant $ q $ learning in spatial action spaces.
\newblock In \emph{Conference on Robot Learning}, pages 1713--1723. PMLR, 2022{\natexlab{b}}.

\bibitem[Wang and Isola(2020)]{wang2020understanding}
Tongzhou Wang and Phillip Isola.
\newblock Understanding contrastive representation learning through alignment and uniformity on the hypersphere.
\newblock In \emph{International conference on machine learning}, pages 9929--9939. PMLR, 2020.

\bibitem[Weiler and Cesa(2019)]{weiler2019general}
Maurice Weiler and Gabriele Cesa.
\newblock General e (2)-equivariant steerable cnns.
\newblock \emph{Advances in neural information processing systems}, 32, 2019.

\bibitem[Worrall et~al.(2017)Worrall, Garbin, Turmukhambetov, and Brostow]{worrall2017harmonic}
Daniel~E Worrall, Stephan~J Garbin, Daniyar Turmukhambetov, and Gabriel~J Brostow.
\newblock Harmonic networks: Deep translation and rotation equivariance.
\newblock In \emph{Proceedings of the IEEE conference on computer vision and pattern recognition}, pages 5028--5037, 2017.

\bibitem[Wu et~al.(2023)Wu, Wen, Li, Li, Yang, and Wang]{wu2023transformation}
Hai Wu, Chenglu Wen, Wei Li, Xin Li, Ruigang Yang, and Cheng Wang.
\newblock Transformation-equivariant 3d object detection for autonomous driving.
\newblock In \emph{Proceedings of the AAAI Conference on Artificial Intelligence}, pages 2795--2802, 2023.

\bibitem[Yu et~al.(2022)Yu, Wu, and Yi]{yu2022rotationally}
Hong-Xing Yu, Jiajun Wu, and Li Yi.
\newblock Rotationally equivariant 3d object detection.
\newblock In \emph{Proceedings of the IEEE/CVF Conference on Computer Vision and Pattern Recognition}, pages 1456--1464, 2022.

\bibitem[Zbontar et~al.(2021)Zbontar, Jing, Misra, LeCun, and Deny]{zbontar2021barlow}
Jure Zbontar, Li Jing, Ishan Misra, Yann LeCun, and St{\'e}phane Deny.
\newblock Barlow twins: Self-supervised learning via redundancy reduction.
\newblock In \emph{International conference on machine learning}, pages 12310--12320. PMLR, 2021.

\end{thebibliography}
